\documentclass[letterpaper, 10 pt, conference]{ieeeconf}  

\IEEEoverridecommandlockouts                              

\usepackage{amsmath} 
\usepackage{amssymb}  
\usepackage{xcolor}
\usepackage{algorithm2e}
\usepackage{graphicx}
\usepackage[font=footnotesize, skip =2pt]{caption}
\usepackage[font=footnotesize]{subcaption}
\usepackage{hyperref}
\usepackage{siunitx}
\usepackage{float}
\usepackage{tikz}
\usepackage{pgfplots}
\usepackage{graphbox}
\usepackage{graphicx}
\usepackage{soul}

\usetikzlibrary{external}

\title{\LARGE \bf
BO-ICP: Initialization of Iterative Closest Point Based on Bayesian Optimization
}

\author{Harel Biggie, Andrew Beathard, and Christoffer Heckman$^*$
\thanks{This work was supported through the DARPA Subterranean Challenge cooperative agreement HR0011-18-2-0043, the National Science Foundation \#1764092, and USDA-NIFA \#2021-67021-33450.}
\thanks{$^{*}$All authors are with the Autonomous Robotics and Perception Group at the
        University of Colorado Boulder, Boulder Colorado, USA. Corresponding author: 
        {\tt\small christoffer.heckman@colorado.edu}}%
}

\begin{document}

\maketitle
\thispagestyle{empty}
\pagestyle{empty}

\begin{abstract}
Typical algorithms for point cloud registration such as Iterative Closest Point (ICP) require a favorable initial transform estimate between two point clouds in order to perform a successful registration. State-of-the-art methods for choosing this starting condition rely on stochastic sampling or global optimization techniques such as branch and bound. In this work, we present a new method based on Bayesian optimization for finding the critical initial ICP transform. We provide three different configurations for our method which highlights the versatility of the algorithm to both find rapid results and refine them in situations where more runtime is available such as offline map building. Experiments are run on popular data sets and we show that our approach outperforms state-of-the-art methods when given similar computation time. Furthermore, it is compatible with other improvements to ICP, as it focuses solely on the selection of an initial transform, a starting point for all ICP-based methods.
\end{abstract}

\section{Introduction} \label{sec:intro}

Point cloud registration or the process of estimating a transformation between two point clouds is paramount to modern robotic applications such as 3D scene reconstruction and decision-making \cite{pomerleau2015review}. The point cloud registration problem is easy to solve if the point correspondences are known. However, in practical applications, the point cloud registration problem is more difficult due to the unknown correspondence of points obtained from sensors such as lidar and stereo cameras. Various registration algorithms to simultaneously estimate the transform and correspondence between two point clouds have been proposed over the last two decades, and Iterative Closest Point (ICP) based methods \cite{besl1992method} have largely dominated the field. ICP-based algorithms aim to minimize the point-to-point correspondence distance between a fixed reference cloud $R$ and a floating target cloud $T$ using an initial transform $T_0$ which in turn provides an estimate of the optimal transform between the two clouds. The results of ICP-based algorithms are highly dependent on the initial transform choice $T_0$. 

At its core, the ICP algorithm works as follows:
\begin{enumerate}
    \item using an initial transform estimate $T_0$, transform the target cloud $T$ into the coordinate frame of reference cloud $R$; 
    \item for each point in target cloud $T$ find the closest corresponding point in the reference cloud $R$;
    \item using an objective (point-to-point distance or similar) find the best transform $T_{i}$ between $T$ and $R$; and
    \item repeat the process until convergence using $T_{i}$ as the the new $T_{0}$.
\end{enumerate}

\begin{figure}[t!]
    \centering
    \includegraphics[width=.48\textwidth]{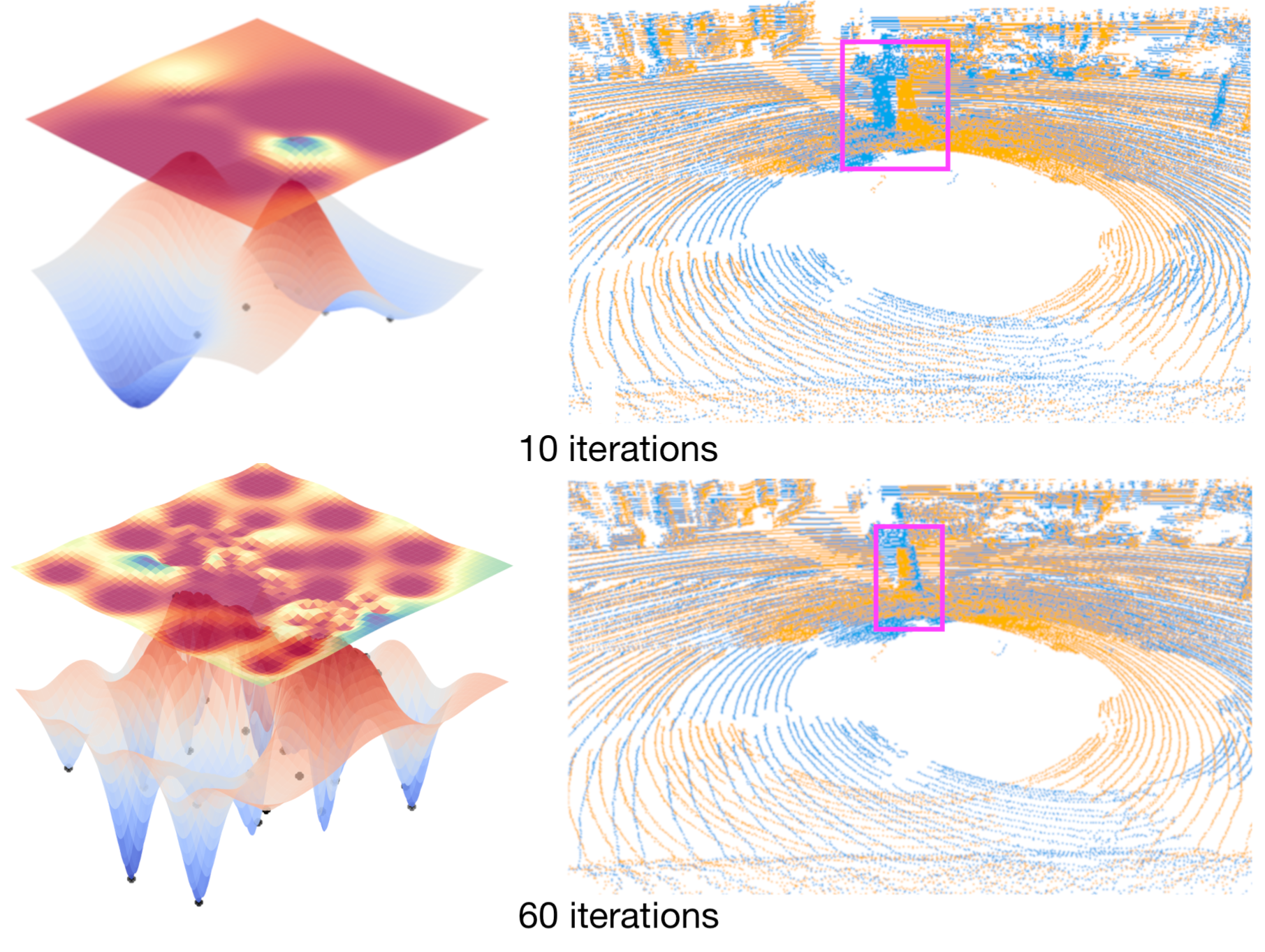}%
    \label{fig:bayes_viz_3d_cloud}
    \caption{A visualization showing a 2D projection of the optimization in BO-ICP on the XY plane. Two different iteration counts of BO-ICP are shown: 10 iterations (top) and 60 iterations (bottom). In each example, a 3D \emph{surrogate function} is shown below a color map of the \emph{acquisition function} where red indicates areas of least utility and blue indicates areas of most utility. Squares highlight key alignment features in the scene.}
    \label{fig:bayes_viz_point}
    \vspace{-7mm}
\end{figure}%

Despite being a highly generalized algorithm, hundreds of variations of the ICP framework exist, e.g, matching points to planes \cite{chen1992object} and colors \cite{park2017colored}. Each of these algorithms was tailored to improve results for data available in specific instances such as RGB information. While many of these algorithms are capable of producing exceptional results given an initial transform estimate $T_{0}$, they are prone to falling into local minima. A poor initial estimate can lead to vastly different results because ICP is an inherently non-convex problem that terminates when the point-to-point correspondence distance stops improving. Finding a globally optimal solution to the ICP problem is an ongoing discipline of research in both the computer vision and robotics communities. 

\begin{figure*}[!t]
     \centering 
     \begin{subfigure}[b]{0.32\textwidth}
         \centering
         \includegraphics[width=\textwidth]{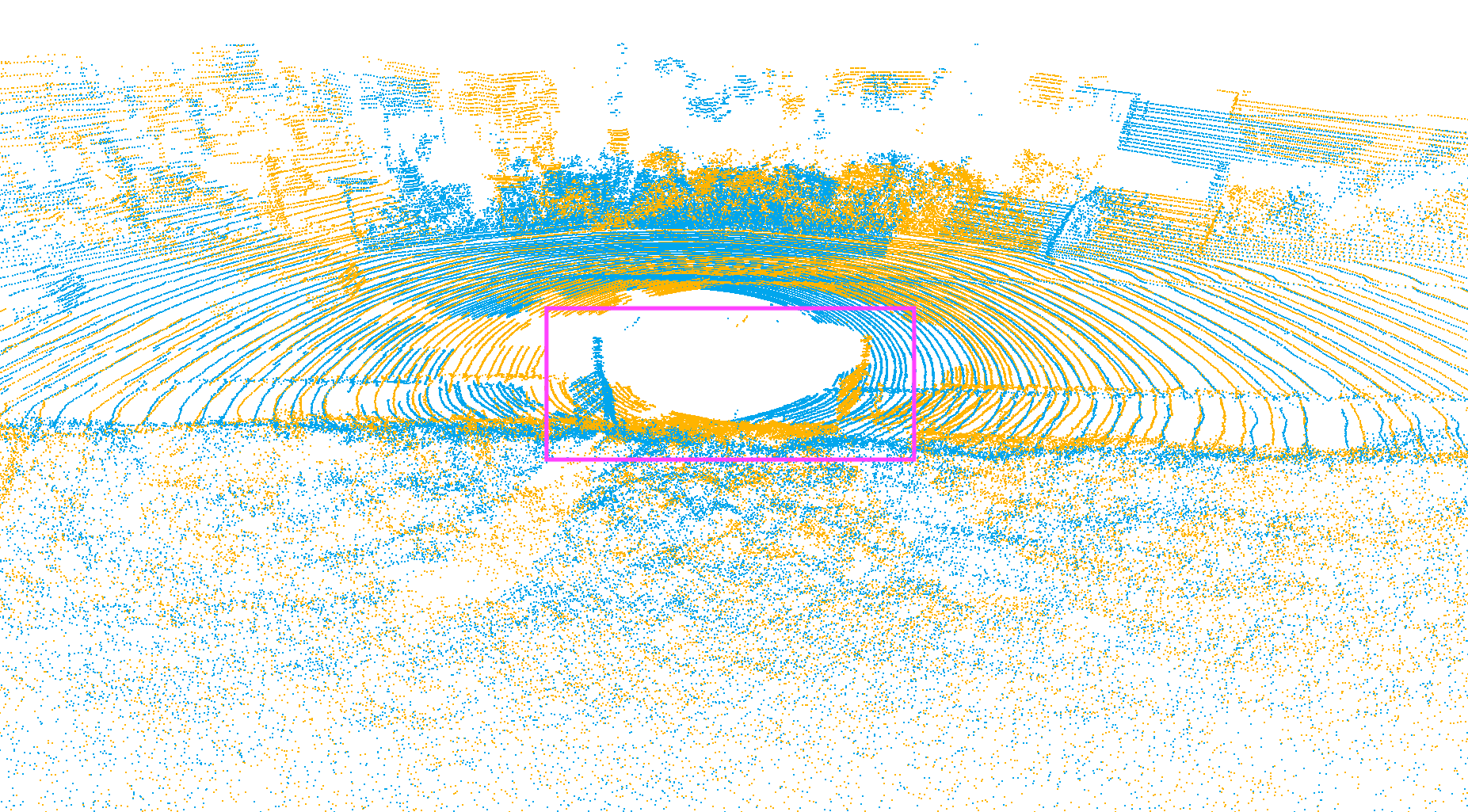}
         \caption{Ground Truth}
         \label{fig:kitti_gt}
     \end{subfigure}
     \hfill
     \begin{subfigure}[b]{0.32\textwidth}
         \centering
         \includegraphics[width=\textwidth]{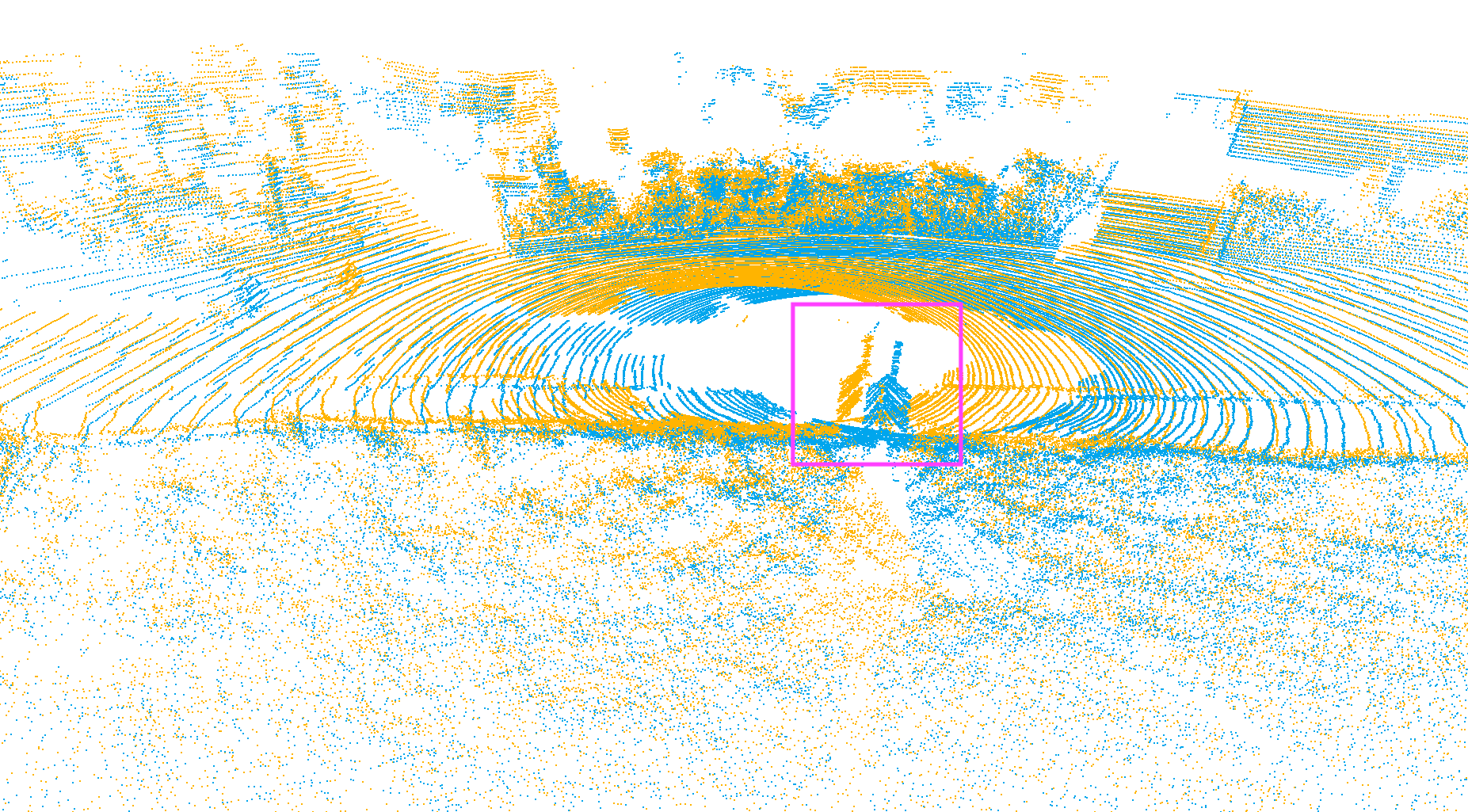}
         \caption{GO-ICP}
         \label{fig:kitti_go_icp}
     \end{subfigure}
     \hfill
     \begin{subfigure}[b]{0.32\textwidth}
         \centering
         \includegraphics[width=\textwidth]{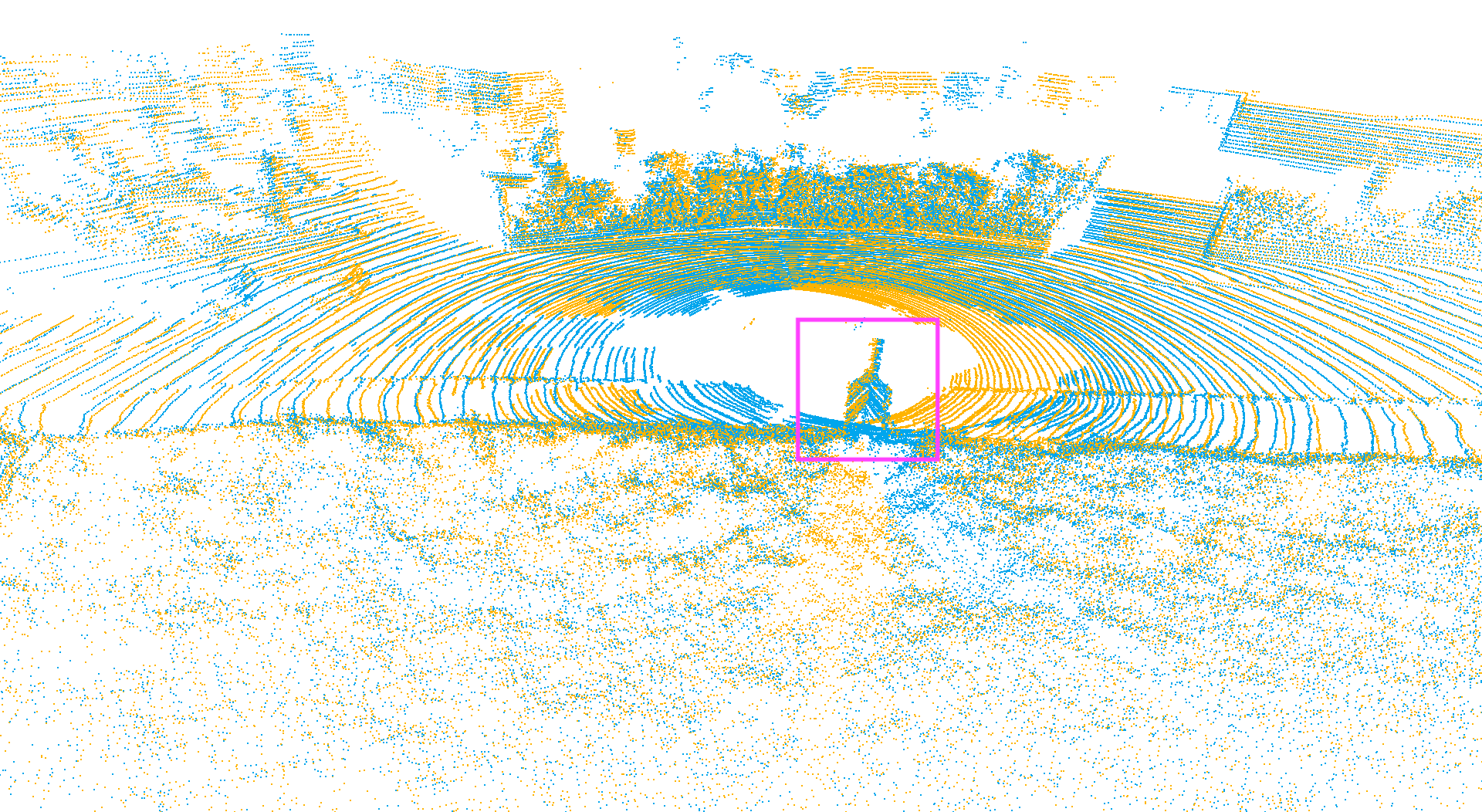}
         \caption{BO-ICP}
         \label{fig:kitti_bayes}
     \end{subfigure}
        \caption{Example alignment of source cloud 000739 (orange) to target cloud 000740 (blue) of sequence 10 from KITTI, with key features highlighted in pink. Due to the inaccuracy of ground truth transforms, GO-ICP and BO-ICP alignment metrics are evaluated on mean point to point distance.}
        \label{fig:kitti_compare}
        \vspace{-7mm}
\end{figure*}


A wide variety of methods have been proposed to either increase the accuracy of $T_0$ through external instrumentation \cite{pomerleau2015review} or the use of feature matching and alignment \cite{zhou2016fast}. Other methods have utilized global optimization techniques such as branch and bound to determine the optimal transform in certain situations such as scan to model matching \cite{yang2013go}. In situations where outliers, dynamic obstacles, or noisy sensor measurements exist, these global methods cannot guarantee an optimal solution \cite{stechschulte2019robust}. The lack of consistently reliable methods for global registration motivates research targeted at more robust initialization techniques. 

In this paper, we outline a framework based on Bayesian optimization (BO) \cite{mockus1978application}, a global optimization method, to systematically compute $T_0$, and find this approach produces more accurate alignments than state-of-the-art methods. It is fundamentally compatible with variants of ICP that address point correspondences, the weighting of correspondences, and other methods that focus on adjusting how the objective is constructed.

Bayesian optimization is a global optimization scheme that has been commonly used to optimize neural network training parameters \cite{snoek2012practical}. In practice, BO aims to find the minima or maxima of an \emph{objective function} and is effective in situations where the objective is either complex, noisy, or expensive to calculate, such as in the case of minimizing the point-to-point correspondence distance. BO utilizes a computationally efficient probabilistic model of the objective function (see Figure \ref{fig:bayes_viz_point}) that is inexpensive to evaluate. As the point cloud registration problem as formulated by ICP is inherently a non-convex problem that requires expensive iterations for each initial $T_0$ estimate, it is an ideal candidate for BO. We demonstrate that our approach outperforms exhaustive searches for this initial estimate, as well as other ``globally optimal'' methods.

In this work, we present an open-source Bayesian optimization-based method (BO-ICP) for determining the crucial initial condition for ICP problems. This method optimizes the initial condition $T_0$ for point cloud registration in a principled manner that is grounded in statistics and is shown to outperform state-of-the-art methods both computationally and in accuracy. Our proposed method is not dependent on prior contextual knowledge. A common approach in such scenarios is to apply random guesses to the initial conditions; we refine this approach through a principled estimation of the objective function, the powerful engine that drives BO. Furthermore, our method generalizes to many of the variants of ICP \cite{pomerleau2015review} as it does not alter the internal operation of the ICP algorithm itself. Finally, our approach is straightforward to integrate into existing point cloud alignment systems, as almost all of them require some initial guess to be provided; our technique provides a principled technique for adjusting this guess.



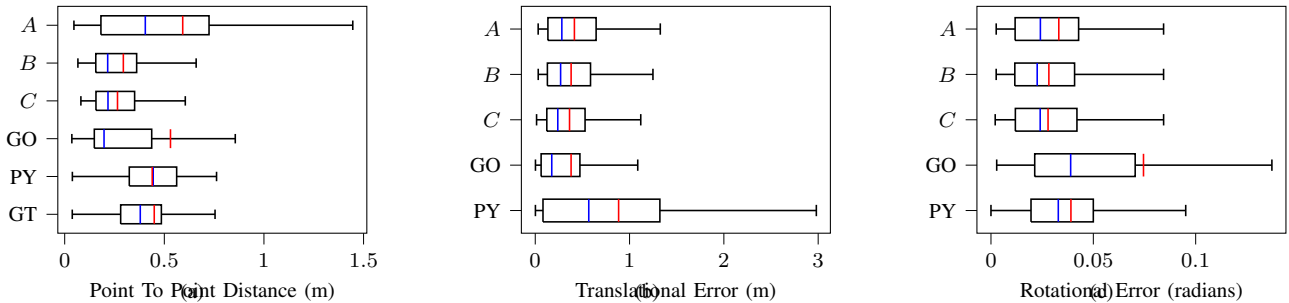
\begin{figure*}[!t]
\vspace{2mm}

     \centering
     \begin{subfigure}[b]{0.32\textwidth}
         \centering
\begin{tikzpicture}

\definecolor{darkgray176}{RGB}{176,176,176}

\begin{axis}[
tick align=outside,
tick pos=left,
x grid style={darkgray176},
xlabel={Point To Point Distance (m)},
xmin=-0.0345491417657814, xmax=1.51635334219223,
xtick style={color=black},
y grid style={darkgray176},
ymin=0.5, ymax=6.5,
ytick style={color=black},
ytick={1,2,3,4,5,6},
yticklabels={GT,PY,GO, $C$,$B$,$A$},
width=\textwidth,
height=46mm,
font=\footnotesize
]
\addplot [semithick, black]
table {%
0.280935780003095 0.75
0.280935780003095 1.25
0.485221055310471 1.25
0.485221055310471 0.75
0.280935780003095 0.75
};
\addplot [semithick, black]
table {%
0.280935780003095 1
0.0380334314277917 1
};
\addplot [semithick, black]
table {%
0.485221055310471 1
0.754597405291726 1
};
\addplot [semithick, black]
table {%
0.0380334314277917 0.875
0.0380334314277917 1.125
};
\addplot [semithick, black]
table {%
0.754597405291726 0.875
0.754597405291726 1.125
};
\addplot [semithick, black]
table {%
0.324194397295417 1.75
0.324194397295417 2.25
0.561989745904076 2.25
0.561989745904076 1.75
0.324194397295417 1.75
};
\addplot [semithick, black]
table {%
0.324194397295417 2
0.0385254089878287 2
};
\addplot [semithick, black]
table {%
0.561989745904076 2
0.762154587993922 2
};
\addplot [semithick, black]
table {%
0.0385254089878287 1.875
0.0385254089878287 2.125
};
\addplot [semithick, black]
table {%
0.762154587993922 1.875
0.762154587993922 2.125
};
\addplot [semithick, black]
table {%
0.148639473062232 2.75
0.148639473062232 3.25
0.436703755556996 3.25
0.436703755556996 2.75
0.148639473062232 2.75
};
\addplot [semithick, black]
table {%
0.148639473062232 3
0.0359464256868556 3
};
\addplot [semithick, black]
table {%
0.436703755556996 3
0.856130829473213 3
};
\addplot [semithick, black]
table {%
0.0359464256868556 2.875
0.0359464256868556 3.125
};
\addplot [semithick, black]
table {%
0.856130829473213 2.875
0.856130829473213 3.125
};
\addplot [semithick, black]
table {%
0.156953731186319 3.75
0.156953731186319 4.25
0.350973143525147 4.25
0.350973143525147 3.75
0.156953731186319 3.75
};
\addplot [semithick, black]
table {%
0.156953731186319 4
0.0813158911640377 4
};
\addplot [semithick, black]
table {%
0.350973143525147 4
0.605270669188664 4
};
\addplot [semithick, black]
table {%
0.0813158911640377 3.875
0.0813158911640377 4.125
};
\addplot [semithick, black]
table {%
0.605270669188664 3.875
0.605270669188664 4.125
};
\addplot [semithick, black]
table {%
0.156063471946464 4.75
0.156063471946464 5.25
0.361170352460553 5.25
0.361170352460553 4.75
0.156063471946464 4.75
};
\addplot [semithick, black]
table {%
0.156063471946464 5
0.0663270307332011 5
};
\addplot [semithick, black]
table {%
0.361170352460553 5
0.659806727559823 5
};
\addplot [semithick, black]
table {%
0.0663270307332011 4.875
0.0663270307332011 5.125
};
\addplot [semithick, black]
table {%
0.659806727559823 4.875
0.659806727559823 5.125
};
\addplot [semithick, black]
table {%
0.181849283371177 5.75
0.181849283371177 6.25
0.724458137330618 6.25
0.724458137330618 5.75
0.181849283371177 5.75
};
\addplot [semithick, black]
table {%
0.181849283371177 6
0.0463450395844797 6
};
\addplot [semithick, black]
table {%
0.724458137330618 6
1.4458577747396 6
};
\addplot [semithick, black]
table {%
0.0463450395844797 5.875
0.0463450395844797 6.125
};
\addplot [semithick, black]
table {%
1.4458577747396 5.875
1.4458577747396 6.125
};
\addplot [semithick, blue]
table {%
0.379341372075546 0.75
0.379341372075546 1.25
};
\addplot [semithick, red]
table {%
0.449448882089208 0.75
0.449448882089208 1.25
};
\addplot [semithick, blue]
table {%
0.444214378393681 1.75
0.444214378393681 2.25
};
\addplot [semithick, red]
table {%
0.439326096093006 1.75
0.439326096093006 2.25
};
\addplot [semithick, blue]
table {%
0.197481250971084 2.75
0.197481250971084 3.25
};
\addplot [semithick, red]
table {%
0.53090807171364 2.75
0.53090807171364 3.25
};
\addplot [semithick, blue]
table {%
0.217304888940132 3.75
0.217304888940132 4.25
};
\addplot [semithick, red]
table {%
0.265031787545264 3.75
0.265031787545264 4.25
};
\addplot [semithick, blue]
table {%
0.215794783204444 4.75
0.215794783204444 5.25
};
\addplot [semithick, red]
table {%
0.29432642442501 4.75
0.29432642442501 5.25
};
\addplot [semithick, blue]
table {%
0.40459389189221 5.75
0.40459389189221 6.25
};
\addplot [semithick, red]
table {%
0.592362416194599 5.75
0.592362416194599 6.25
};
\end{axis}

\end{tikzpicture}
         \vspace*{-10mm}
         \caption{}
         \label{fig:kitti_box}
     \end{subfigure}%
     \hfill
     \begin{subfigure}[b]{0.32\textwidth}
         \centering
\begin{tikzpicture}

\definecolor{darkgray176}{RGB}{176,176,176}

\begin{axis}[
tick align=outside,
tick pos=left,
x grid style={darkgray176},
xlabel={Translational Error (m)},
xmin=-0.147169809632545, xmax=3.12953130825477,
xtick style={color=black},
y grid style={darkgray176},
ymin=0.5, ymax=5.5,
ytick style={color=black},
ytick={1,2,3,4,5},
yticklabels={PY,GO,$C$,$B$,$A$},
width=\textwidth,
height=46mm,
font=\footnotesize
]
\addplot [semithick, black]
table {%
0.0820853752768552 0.75
0.0820853752768552 1.25
1.32156047875692 1.25
1.32156047875692 0.75
0.0820853752768552 0.75
};
\addplot [semithick, black]
table {%
0.0820853752768552 1
0.00177115027142429 1
};
\addplot [semithick, black]
table {%
1.32156047875692 1
2.9805903483508 1
};
\addplot [semithick, black]
table {%
0.00177115027142429 0.875
0.00177115027142429 1.125
};
\addplot [semithick, black]
table {%
2.9805903483508 0.875
2.9805903483508 1.125
};
\addplot [semithick, black]
table {%
0.0621964968326653 1.75
0.0621964968326653 2.25
0.473588554450781 2.25
0.473588554450781 1.75
0.0621964968326653 1.75
};
\addplot [semithick, black]
table {%
0.0621964968326653 2
0.00224014004135667 2
};
\addplot [semithick, black]
table {%
0.473588554450781 2
1.08680631373935 2
};
\addplot [semithick, black]
table {%
0.00224014004135667 1.875
0.00224014004135667 2.125
};
\addplot [semithick, black]
table {%
1.08680631373935 1.875
1.08680631373935 2.125
};
\addplot [semithick, black]
table {%
0.122640289894179 2.75
0.122640289894179 3.25
0.528192071095523 3.25
0.528192071095523 2.75
0.122640289894179 2.75
};
\addplot [semithick, black]
table {%
0.122640289894179 3
0.0140309639437255 3
};
\addplot [semithick, black]
table {%
0.528192071095523 3
1.11878296272017 3
};
\addplot [semithick, black]
table {%
0.0140309639437255 2.875
0.0140309639437255 3.125
};
\addplot [semithick, black]
table {%
1.11878296272017 2.875
1.11878296272017 3.125
};
\addplot [semithick, black]
table {%
0.128701871493625 3.75
0.128701871493625 4.25
0.586075850503371 4.25
0.586075850503371 3.75
0.128701871493625 3.75
};
\addplot [semithick, black]
table {%
0.128701871493625 4
0.0316007125039247 4
};
\addplot [semithick, black]
table {%
0.586075850503371 4
1.24798505924837 4
};
\addplot [semithick, black]
table {%
0.0316007125039247 3.875
0.0316007125039247 4.125
};
\addplot [semithick, black]
table {%
1.24798505924837 3.875
1.24798505924837 4.125
};
\addplot [semithick, black]
table {%
0.134266974564683 4.75
0.134266974564683 5.25
0.645433509642281 5.25
0.645433509642281 4.75
0.134266974564683 4.75
};
\addplot [semithick, black]
table {%
0.134266974564683 5
0.0309729960795458 5
};
\addplot [semithick, black]
table {%
0.645433509642281 5
1.32589702076082 5
};
\addplot [semithick, black]
table {%
0.0309729960795458 4.875
0.0309729960795458 5.125
};
\addplot [semithick, black]
table {%
1.32589702076082 4.875
1.32589702076082 5.125
};
\addplot [semithick, blue]
table {%
0.567468877115827 0.75
0.567468877115827 1.25
};
\addplot [semithick, red]
table {%
0.883970700304634 0.75
0.883970700304634 1.25
};
\addplot [semithick, blue]
table {%
0.175059842497605 1.75
0.175059842497605 2.25
};
\addplot [semithick, red]
table {%
0.379998318914375 1.75
0.379998318914375 2.25
};
\addplot [semithick, blue]
table {%
0.238988984254134 2.75
0.238988984254134 3.25
};
\addplot [semithick, red]
table {%
0.363531697227215 2.75
0.363531697227215 3.25
};
\addplot [semithick, blue]
table {%
0.268911989607337 3.75
0.268911989607337 4.25
};
\addplot [semithick, red]
table {%
0.380344847770828 3.75
0.380344847770828 4.25
};
\addplot [semithick, blue]
table {%
0.281985684980812 4.75
0.281985684980812 5.25
};
\addplot [semithick, red]
table {%
0.415137321164127 4.75
0.415137321164127 5.25
};
\end{axis}

\end{tikzpicture}
         \vspace*{-10mm}
         \caption{}
         \label{fig:tum_trans_error}
     \end{subfigure}%
     \hfill
     \begin{subfigure}[b]{0.32\textwidth}
         \centering
\begin{tikzpicture}

\definecolor{darkgray176}{RGB}{176,176,176}

\begin{axis}[
tick align=outside,
tick pos=left,
x grid style={darkgray176},
xlabel={Rotational Error (radians)},
xmin=-0.00686202150505854, xmax=0.144102451606229,
xtick style={color=black},
y grid style={darkgray176},
ymin=0.5, ymax=5.5,
ytick style={color=black},
ytick={1,2,3,4,5},
yticklabels={PY,GO,$C$,$B$,$A$},
width=\textwidth,
height=46mm,
font=\footnotesize,
xticklabel style={
  /pgf/number format/precision=3,
  /pgf/number format/fixed}
]
\addplot [semithick, black]
table {%
0.0195609763331729 0.75
0.0195609763331729 1.25
0.0499958505858523 1.25
0.0499958505858523 0.75
0.0195609763331729 0.75
};
\addplot [semithick, black]
table {%
0.0195609763331729 1
0 1
};
\addplot [semithick, black]
table {%
0.0499958505858523 1
0.0951375821431748 1
};
\addplot [semithick, black]
table {%
0 0.875
0 1.125
};
\addplot [semithick, black]
table {%
0.0951375821431748 0.875
0.0951375821431748 1.125
};
\addplot [semithick, black]
table {%
0.0213713685774379 1.75
0.0213713685774379 2.25
0.0704501134625814 2.25
0.0704501134625814 1.75
0.0213713685774379 1.75
};
\addplot [semithick, black]
table {%
0.0213713685774379 2
0.00279968314769634 2
};
\addplot [semithick, black]
table {%
0.0704501134625814 2
0.137240430101171 2
};
\addplot [semithick, black]
table {%
0.00279968314769634 1.875
0.00279968314769634 2.125
};
\addplot [semithick, black]
table {%
0.137240430101171 1.875
0.137240430101171 2.125
};
\addplot [semithick, black]
table {%
0.0119126604505618 2.75
0.0119126604505618 3.25
0.0419968152047658 3.25
0.0419968152047658 2.75
0.0119126604505618 2.75
};
\addplot [semithick, black]
table {%
0.0119126604505618 3
0.00206680583236349 3
};
\addplot [semithick, black]
table {%
0.0419968152047658 3
0.0843389764453648 3
};
\addplot [semithick, black]
table {%
0.00206680583236349 2.875
0.00206680583236349 3.125
};
\addplot [semithick, black]
table {%
0.0843389764453648 2.875
0.0843389764453648 3.125
};
\addplot [semithick, black]
table {%
0.0116481452127315 3.75
0.0116481452127315 4.25
0.0408441796849194 4.25
0.0408441796849194 3.75
0.0116481452127315 3.75
};
\addplot [semithick, black]
table {%
0.0116481452127315 4
0.00254689618110893 4
};
\addplot [semithick, black]
table {%
0.0408441796849194 4
0.0843263811106218 4
};
\addplot [semithick, black]
table {%
0.00254689618110893 3.875
0.00254689618110893 4.125
};
\addplot [semithick, black]
table {%
0.0843263811106218 3.875
0.0843263811106218 4.125
};
\addplot [semithick, black]
table {%
0.0117942428855066 4.75
0.0117942428855066 5.25
0.0428146695156243 5.25
0.0428146695156243 4.75
0.0117942428855066 4.75
};
\addplot [semithick, black]
table {%
0.0117942428855066 5
0.0025576482793745 5
};
\addplot [semithick, black]
table {%
0.0428146695156243 5
0.0843263811106218 5
};
\addplot [semithick, black]
table {%
0.0025576482793745 4.875
0.0025576482793745 5.125
};
\addplot [semithick, black]
table {%
0.0843263811106218 4.875
0.0843263811106218 5.125
};
\addplot [semithick, blue]
table {%
0.0328903451133339 0.75
0.0328903451133339 1.25
};
\addplot [semithick, red]
table {%
0.0390599677917558 0.75
0.0390599677917558 1.25
};
\addplot [semithick, blue]
table {%
0.0389283990723497 1.75
0.0389283990723497 2.25
};
\addplot [semithick, red]
table {%
0.0745116845005994 1.75
0.0745116845005994 2.25
};
\addplot [semithick, blue]
table {%
0.0240423049265749 2.75
0.0240423049265749 3.25
};
\addplot [semithick, red]
table {%
0.0278856971703646 2.75
0.0278856971703646 3.25
};
\addplot [semithick, blue]
table {%
0.0225810164041075 3.75
0.0225810164041075 4.25
};
\addplot [semithick, red]
table {%
0.0282845766515904 3.75
0.0282845766515904 4.25
};
\addplot [semithick, blue]
table {%
0.0241385154194884 4.75
0.0241385154194884 5.25
};
\addplot [semithick, red]
table {%
0.0331344477796242 4.75
0.0331344477796242 5.25
};
\end{axis}

\end{tikzpicture}%
         \vspace*{-10mm}
         \caption{}
         \label{fig:tum_rot_error}
     \end{subfigure}%
      \caption{Statistical summaries for the point to point distance in KITTI (a), translational error (b), rotational error (c) in TUM. The vertical box edges signify the first and third quartiles; the whiskers extend from the box edges to the minimum/maximum value in either direction, up to a max distance of 1.5 $\times$ the interquartile range, outliers are not shown. Means are shown in red and medians are shown in blue. GT represents ground truth for the KITTI datasets and $A$ - $E$ are the different configurations for BO-ICP. }
      \vspace{-7mm}
        \label{fig:box_plots}
\end{figure*}

\section{RELATED WORKS}

An abundance of high-quality prior work has been produced on point cloud registration. As our method directly contributes to initialization procedures for ICP and global point cloud registration, we will focus our review on this facet of the field and refer the reader to the following surveys for a more broad view of the field \cite{pomerleau2015review, tam2012registration, pomerleau2013comparing}.

\subsubsection*{Initial Transform Estimation}

A wide variety of methods generally falling into three categories have been developed for estimating the required $T_0$ in ICP methods \cite{pomerleau2015review}. \emph{External} based methods rely on obtaining the estimate from an outside source such as a human in the loop \cite{godin1994three}, Inertial Measurement Unit (IMU), or GPS \cite{diebel2004simultaneous, druon2006color, gao2015ins}. Relying on external sensors limits the application space of these registration methods and prevents the generalization of various solutions. \emph{Parameter cascade} based methods generally modify parameters such as the resolution of the point cloud through multiple iterations in order to achieve faster results. These methods commonly operate on a coarse-to-fine strategy where computationally results are obtained by down-sampling the point cloud and later refined with a better initial guess \cite{jost2002multi} or by sectioning off the point cloud into local regions \cite{bosse2009continuous}. \emph{Registration Cascade} based methods are a class of initialization methods that rely on a separate external process to estimate the initial transformation. These approaches generally use histograms \cite{bosse2008map, rusu2009fast} or keyframe-based approaches \cite{tsai2009edge} to determine an initial $T_0$.

\subsubsection*{Optimization Based Approaches}
Other approaches have attempted to use optimization techniques such as the Levenberg–Marquardt algorithm \cite{more1978levenberg} to perform point cloud registration \cite{fitzgibbon2003robust}. The authors found that in certain situations the algorithm converged faster than traditional ICP based methods. Gaussian mixture models (GMMs) have also widely been explored to locally refine the initial transform \cite{jian2005robust, campbell2015adaptive}. While these methods are capable of optimizing the estimate locally, they still require an initialization for $T_0$.

\subsubsection*{Global Methods}
More recent methods have used global optimization techniques such as branch and bound to solve the 3D registration problem \cite{enqvist2009optimal, yang2013go, yang2015go}. A mathematical guarantee of convergence to the optimal solution was presented in GO-ICP \cite{yang2013go} using a nested branch and bound approach. GO-ICP computes inner and outer bounds for both the $\mathbb{SO}(3)$ rotation and $\mathbb{SE}(3)$ translation spaces using an uncertainty radius approach. The method works remarkably well for model to scan alignments and small object to scene matching. However, it requires that both input clouds are scaled into the same unit cube and is not robust to low overlap conditions \cite{stechschulte2019robust}. \cite{choi2015robust} presents a RANSAC-based geometric feature extraction method that is used to compute the correspondence inside of a fast point matching histogram. This method has the potential for long run times due to the need to evaluate a potential infinite number of feature extractions. Fast Global Registration \cite{zhou2016fast} overcomes some of these limitations using a combination of Fast Point Feature Histograms and an efficient optimization approach based on Black-Rangarajan duality \cite{black1996unification}. Algorithms based on 4PCS \cite{aiger20084, mellado2014super} perform global registration by utilizing intelligent selections of congruent points. More recently, \cite{yang2020teaser} presents a truncated least squares approach combined with a maximal clique inlier selection procedure to solve the global registration problem in a certifiable manner that is robust to outliers.




\begin{figure*}[tbp]
     \centering
     \begin{subfigure}[b]{0.32\textwidth}
         \centering
         \includegraphics[width=\textwidth]{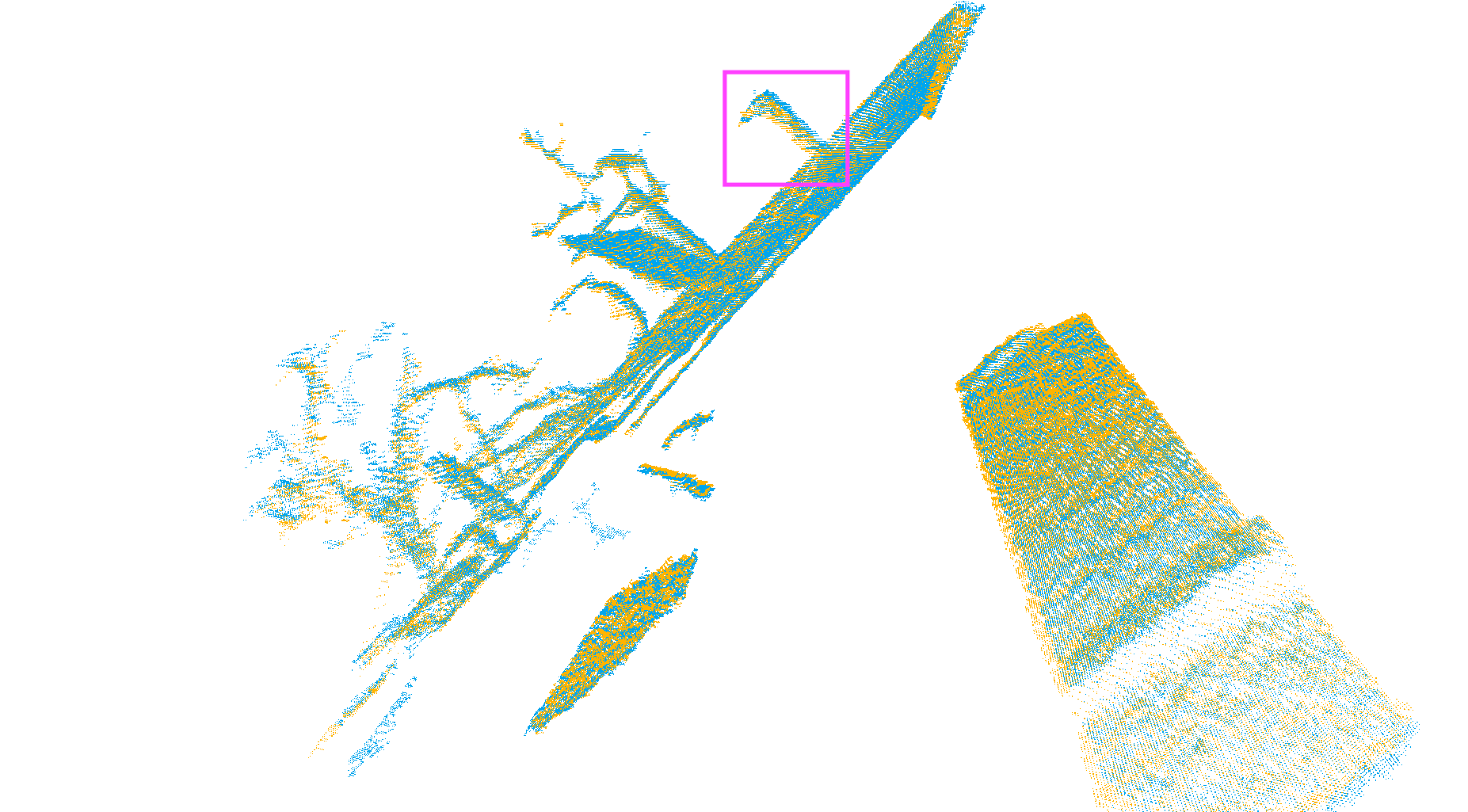}
         \caption{Ground Truth}
         \label{fig:tum_gt}
     \end{subfigure}
     \hfill
     \begin{subfigure}[b]{0.32\textwidth}
         \centering
         \includegraphics[width=\textwidth]{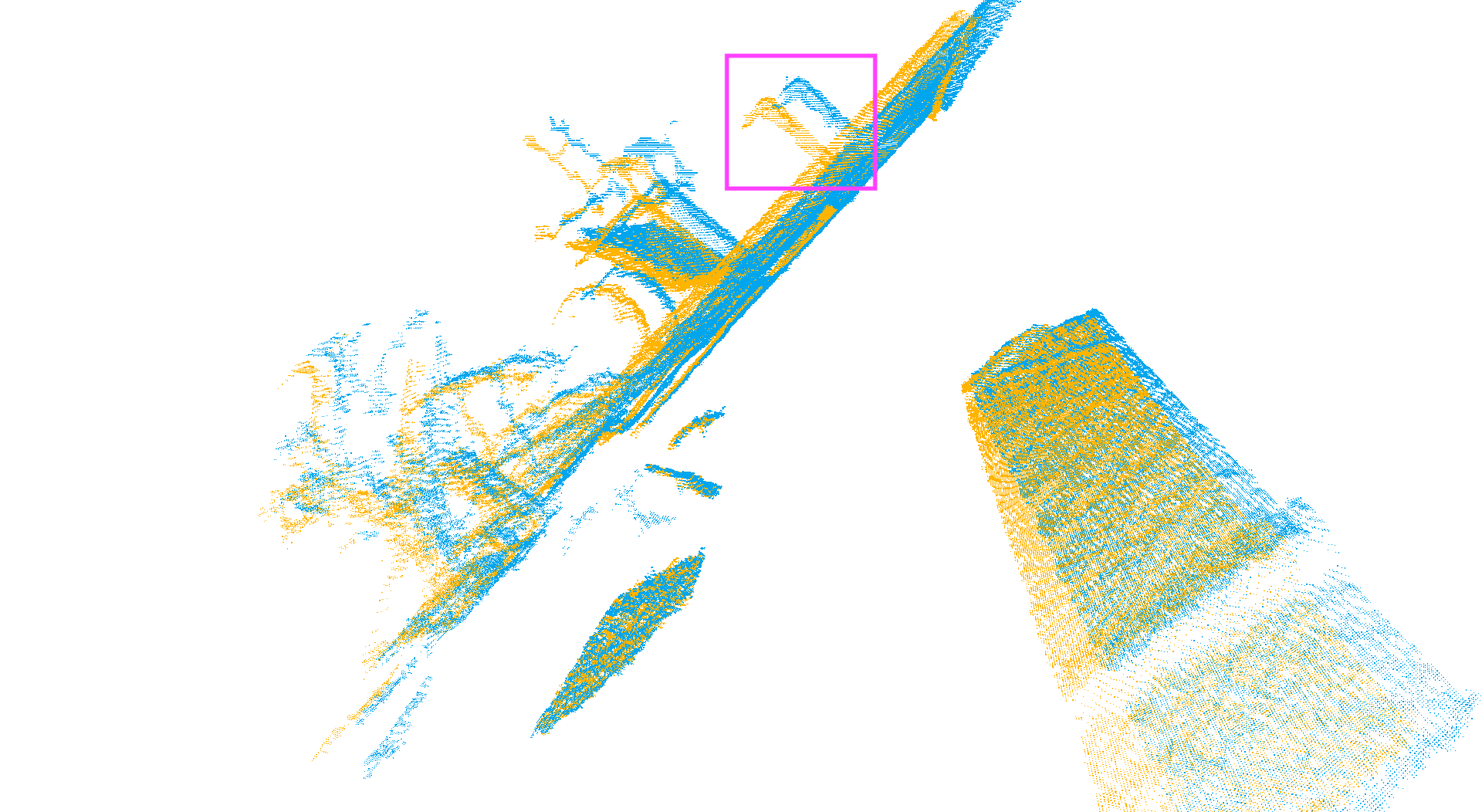}
         \caption{GO-ICP}
         \label{fig:tum_go_icp}
     \end{subfigure}
     \hfill
     \begin{subfigure}[b]{0.32\textwidth}
         \centering
         \includegraphics[width=\textwidth]{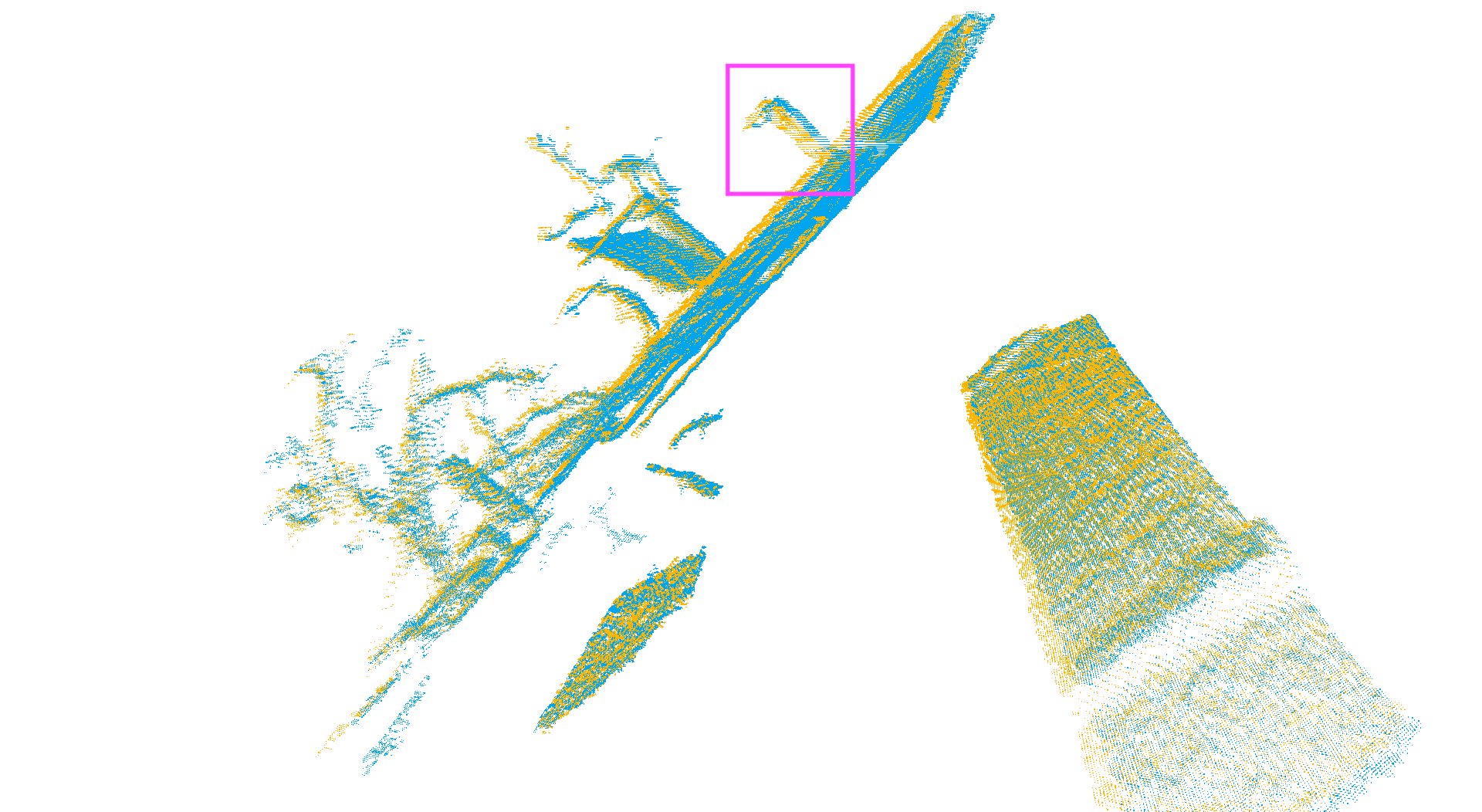}
         \caption{BO-ICP}
         \label{fig:tum_bayes}
     \end{subfigure}
      \caption{Example alignment of source cloud timestamp 1311868250.071665 (orange) to target cloud timestamp 1311868250.142995 (blue) of sequence fr2/desk of the TUM odometry dataset. Pink squares highlight key alignment features.}
        \label{fig:tum_compare}
    \vspace{-7mm}
\end{figure*}

\section{Methodology}
Consider the problem of aligning a target point cloud $S$ to a reference cloud $R$ through a rigid geometric transformation $T \in \mathbb{SE}(3)$, which we parametrize by $[x,y,z, \theta, \psi, \phi]$, with $\theta$, $\psi$, and $\phi$ representing roll, pitch, and yaw respectively. In ICP, this is accomplished through the minimization of an \emph{objective function} which is calculated over the two point sets; see step three of the ICP algorithm described in Section \ref{sec:intro}. A challenge here is that the initial condition to ICP $T_0$ is arbitrary, and has a significant impact on the convergence and results of the final alignment. In order to address this, we develop a technique here that principally identifies $T_g$, a global transform for use as $T_0$ in ICP, using BO a global optimization technique. Note that BO can fundamentally be applied to any objective function used inside of ICP, e.g., point-to-point \cite{besl1992method}, point-to-plane \cite{chen1992object}, color \cite{park2017colored}; we focus on its application towards optimizing the $T_0$ for the point-to-point ICP  \cite{besl1992method} for this discussion.

\setlength{\textfloatsep}{0pt}

\RestyleAlgo{ruled}

\begin{algorithm}[htpb]
\caption{BO-ICP: For full 6DOF search. The algorithm can be done with rotations $[\theta, \psi, \phi]$ and then translations $[x,y,z]$. Where the identity is used as the initial translation and the optimized rotation is used to find the optimized translation.}\label{alg:bo_icp}
\KwData{Source cloud $\mathbf{S}$, Reference cloud $\mathbf{R}$, bounds $\mathbf{b}$, Number of random steps $\mathbf{N_r}$, Number of iterations $\mathbf{N}$}
\KwResult{Transformation between the two clouds $\mathbf{T_f}$}
 Populate $\mu(T)$ for $N_r$ random $T_i$ generated from  $[x_i, y_i, z_i, \theta_i, \psi_i, \phi_i]$ within bounds $b$\;
 \While{$i$ $<$ $N$}{
  $T_i \leftarrow$ max utility $[x_i, y_i, z_i, \theta_i, \psi_i, \phi_i]$ from $EI(x)$\;
  $E(T_i) \leftarrow$ Point To Point ICP($T_i$, $S$, $R$)\;
  Update $\mu(T_i)$ with $E(T_i)$\;
 }
 $T_g \leftarrow$ at min $E(T_i)$;
\end{algorithm}%
\setlength{\textfloatsep}{0pt}

\subsubsection*{Point-To-Point ICP}

For the point-to-point formulation of ICP, let $S$ represent the source cloud that is being aligned to the reference cloud $R$ using a transformation $T$. The algorithm works by finding the set of corresponding points $K$ between each of the point clouds $S$ and $R$ transformed by $T$, $K = \{(s,r)\}$. Next, the algorithm aims to minimize an objective function representing the point-to-point correspondence distance between the points (Equation \ref{eq:p2pobjective}).

\begin{equation} \label{eq:p2pobjective}
    E(T_i)  = \sum_{(s,r)} ||r -T_i s||^{2}
\end{equation}

\noindent The process is repeated with $T_i$ becoming the newly minimized transform obtained from each iteration $i$. This iterative process is highly susceptible to converging to a local minimum which can result in erroneous alignments, especially with a poor initial alignment. However, trying all possible alignments is largely considered an NP-hard problem \cite{enqvist2009optimal} and which motivates the need for a more computationally efficient approach.

\subsubsection*{Bayesian Optimization}

The crux of BO is an efficient approximation based on Bayes Rule of the objective function; this is known as the \emph{surrogate function} which we call $\mu(T)$. Next, an \emph{acquisition function} is used to select the next best sample from the posterior search space modeled by the $\mu(T)$. Performing global optimization on ICP based methods is an ideal use case for BO due to the inherent non-convex nature of the functions and the expensive computations required at each iteration. By estimating the ICP function using a \emph{surrogate function} BO is able to efficiently choose the next best guess to minimize the uncertainty based on the \emph{acquisition function}, in comparison to evaluating $E(T_i)$ at each iteration. 

Typically the \emph{surrogate function} is modeled using Gaussian processes which can be used to directly infer a distribution over a ``black box'' function. Gaussian processes use prior observations to learn a posterior over a given function. We make an assumption here that the ICP function can be modeled as a random variable distributed over a Gaussian.

Once the \emph{surrogate function} is modeled one can evaluate the utility of sampling at a new point using a variety of probabilistic models. For the purposes of BO-ICP we use the expected improvement (EI) metric. Let $y^*$ represent the minimum value observed from the objective function, $x$ represent the modeled parameter, and $y$ represent the possible new observations. Then $p(y|x)$ is representative of the surrogate model which is modeled as a Gaussian. The EI can then be written as:

\begin{equation}
    EI_{y^*}(x) = \int_{-\infty}^\infty \max(y^* -y, 0)p(y|x)dy
\end{equation}

BO-ICP can be run in two variations, a full six degrees of freedom (6DOF) search in $T$ (shown in Algorithm \ref{alg:bo_icp}), and a more practical nested approach for computational efficiency. In all cases, the bounds $b$ of the problem should be chosen based on the range of the sensor being used and the size of the scenes. As shown in \cite{yang2013go} it is possible to first search for the optimal rotation parameters ($[\theta, \psi, \phi])$ and then the translation parameters $[x,y,z]$. While we note BO does not share the same theoretical guarantee behind this nested approach we find that in practice we are able to leverage this technique to help reduce the number of hyperparameters required by the \emph{surrogate function}. In the nested variant, we first run BO on only the rotation parameters, and then we pass the computed rotation that minimizes the objective through BO again to find the optimal translation. For both the translation and rotation searches we sample a pseudo-random set of points to build an initial estimate of the \emph{surrogate function}. We note that generally, sampling in $[\theta, \psi, \phi])$ is susceptible to gimbal lock which we overcome by centering our axis around the sample and applying a change of basis. Sampling in terms of quaternions was also explored but the conversion of sampled points to a uniform quaternion as described in \cite{shoemake19926} breaks the normal assumption present in the GP. In practice, this is not strictly required but greatly aids in the ability of the \emph{acquisition function} to make initially informed guesses. 
 
Determining the termination criteria for BO-ICP is difficult in practice without knowing any prior information about the sensor. ICP will locally terminate when the point-to-point distance becomes stationary. Given enough iterations, BO will also reach a point where the objective stops improving but in practice, the optimal value will be found long before this occurs. As such, we select $T_0$ which best minimizes $E(T)$ after running the algorithm for a set number of iterations $N$. Additionally, as described in Section \ref{sec:experiments} we conduct experiments to find values for $N$ which work well in practice.

\section{Experiments}
\label{sec:experiments}

We compare the nested variant of BO-ICP against two other global methods: a brute-force pyramid-based grid search (PY) and GO-ICP \cite{yang2013go} which, to the best of our knowledge, is the most successful global ICP variant that does not rely on selecting informative geometries or feature point extraction. Experiments are conducted from the KITTI Odometry dataset \cite{Geiger2012CVPR, Geiger2013IJRR} and the TUM Vision Handheld SLAM RGB-D dataset \cite{sturm12iros} which enables us to test the robustness of our method to realistic noise. The KITTI dataset was chosen to highlight the effectiveness of our method in the presence of fast-moving scenes and dynamic obstacles. The TUM dataset was collected using a structured light sensor which allows us to evaluate our dataset on point clouds with more noise in comparison to the lidar-based clouds of the KITTI dataset. Unlike the inaccurate GPS-based ground truth in the KITTI dataset (see Section \ref{sec:results} for more), the TUM dataset provides ground truth obtained from a motion capture system which allows us to conduct an accurate analysis of the absolute error in transformations.

Pyramid-based searches are one way of addressing local minima in ICP. In our comparisons, we follow a nested approach where we first search for rotation and then translation. An initial grid of size of 12$\times$12$\times$12 (1728 samples) is populated using a radius-based nearest neighbor search to estimate the fit of the initial transform. New grids of size 6$\times$6$\times$6 are then created from the top 10 results using a fixed delta value from the parameter and the process is repeated for a total of 3 times. At completion, ICP is performed on the top 10 values and the top result is selected. The resulting rotation values from the first search are used to find the optimal translational component using the same procedure.

Intervals of testing data are selected from the KITTI and TUM datasets using a pseudo-random initialization procedure. From the selected intervals, reference $R$  and target $T$ point clouds are paired together for testing using a conservative cloud overlap estimation from \cite{stechschulte2019robust}. We use this metric to ensure there is sufficient overlap between the two clouds (70\% on KITTI and 60\% on TUM) for ICP to be performed. Once that threshold is surpassed, the target cloud $T$ becomes the new reference cloud $R$. This method produces 248 testing pairs from the KITTI dataset and 325 testing pairs from the TUM dataset on which we analyze the efficacy of the proposed method.

\begin{table}[!htb]
\centering
\begin{tabular}{|c|c|c|c|}
\hline
\textbf{Config} & \textbf{Random Samples} & \textbf{Iterations} & \textbf{Voxel Size}     \\ \hline
$A$    & 10                      & 20                  & 0.7                                                    \\ \hline
$B$    & 20                      & 35                  & 0.7                                                       \\ \hline
$C$    & 30                      & 60                  & 0.6                                                     \\ \hline
\end{tabular}
\caption{Summary of the testing configurations for BO-ICP. Configuration $A$ is designed to emphasize the ability of the algorithm to quickly find reasonable alignments while $B$ serves as a direct comparison to GO-ICP, and $C$ shows that BO-ICP is capable of further refining measurements for map building.}
\label{table:testing_config}
\end{table}

For each alignment pair, we select $r$ pseudo-random initialization points and then run Bayesian Optimization for $N$ iterations to find the best rotation estimate. We select 3 different configurations for BO-ICP which highlight the algorithm's versatility: 

\begin{itemize}
    \item Configuration $A$ illustrates the potential of the method to run as part of a loop closure mechanism in the context of SLAM or when realigning maps in the online navigational setting.
    \item Configuration $B$ serves as a comparison between BO-ICP, GO-ICP, and the pyramid algorithm when allowed to run for similar durations. 
    \item Configuration $C$ emphasizes BO-ICP's ability to continuously improve estimates when given more computation time which can be utilized in batch processing large offline maps.
\end{itemize}

  Parameters were chosen to show trends between the different methodologies rather than directly comparing their implementations. We note that ICP implementation in GO-ICP is single-threaded while the other methods are multithreaded. However, on profiling, the code the bulk of the runtime occurs in the generation of the distance transformation required for the algorithm. For each alignment pair, we optimize over the full rotation space and the following bounds for translation: $x\in [-4,4]$, $y \in [-2,2]$, $z \in [-1,1]$. The bounds were selected based on the velocity of the vehicle in the KITTI dataset. 

Summaries of the important parameters for each configuration can be found in Table \ref{table:testing_config} including the voxel-based down-sampling approach used for efficiency. We note that the voxel size and the number of iterations have the most significant impact on the performance of BO-ICP. Other parameters required minimal adjustments and the full parameter set as well as our open source implementation can be found at \url{https://github.com/arpg/bo_icp}. The provided implementation also provides configurations for the full 6DOF variant of BO-ICP. However, experiments for this variant showed that the full version of BO-ICP quadrupled the runtime without any noticeable improvement in results.



Our Python implementation leverages the Open3D Point Cloud Library \cite{Zhou2018} and an open source Bayesian optimization package \cite{bayeslibrary}. Our implementation utilizes multi-threading for ICP but not in the BO process. We compare our implementation to the C++ implementation of GO-ICP provided by the authors \cite{yang2013go} using the default configuration except for a trim fraction of 0.3 to prevent failures from outliers in the data \cite{chetverikov2002trimmed}. All presented experimental results were run on an Intel i7-8850H CPU with 32GB of RAM.

\section{Results}
\label{sec:results}




\begin{table}[htb]
\begin{tabular}{c|cccccc|}
\cline{2-7}
\textbf{}                         & \multicolumn{4}{c|}{\textbf{Runtime}}                                                                                                 & \multicolumn{2}{c|}{\textbf{Error}}                                     \\ \cline{2-7} 
                                  & \multicolumn{2}{c|}{\textit{\textbf{KITTI}}}                        & \multicolumn{4}{c|}{\textit{\textbf{TUM}}}                                                                                                    \\ \cline{2-7} 
                                  & \multicolumn{1}{c|}{$\mu$ \textit{(s)}} & \multicolumn{1}{c|}{$\sigma$ \textit{(s)}} & \multicolumn{1}{c|}{$\mu$ \textit{(s)}} & \multicolumn{1}{c|}{$\sigma$ \textit{(s)}} & \multicolumn{1}{c|}{\textit{Trans (m)}} & \textit{Rot (rad)} \\ \hline
\multicolumn{1}{|c|}{\textit{A}}  & \multicolumn{1}{c|}{\textbf{11.3}}            & \multicolumn{1}{c|}{1.3}            & \multicolumn{1}{c|}{\textbf{12.3}}            & \multicolumn{1}{c|}{\textbf{0.8}}            & \multicolumn{1}{c|}{0.42}                         &       0.033                  \\ \hline
\multicolumn{1}{|c|}{\textit{B}}  & \multicolumn{1}{c|}{22.3}            & \multicolumn{1}{c|}{2.0}            & \multicolumn{1}{c|}{22.5}            & \multicolumn{1}{c|}{1.7}            & \multicolumn{1}{c|}{0.38}                         &       0.028                  \\ \hline
\multicolumn{1}{|c|}{\textit{C}}  & \multicolumn{1}{c|}{43.1}            & \multicolumn{1}{c|}{2.3}            & \multicolumn{1}{c|}{40.0}            & \multicolumn{1}{c|}{3.7}            & \multicolumn{1}{c|}{\textbf{0.36}}                         &       \textbf{0.028}                  \\ \hline
\multicolumn{1}{|c|}{\textit{GO}} & \multicolumn{1}{c|}{25.1}            & \multicolumn{1}{c|}{\textbf{0.6}}            & \multicolumn{1}{c|}{26.0}            & \multicolumn{1}{c|}{0.9}            & \multicolumn{1}{c|}{0.37}                         &       0.074                  \\ \hline
\multicolumn{1}{|c|}{\textit{PY}} & \multicolumn{1}{c|}{28.4}            & \multicolumn{1}{c|}{6}            & \multicolumn{1}{c|}{11.4}            & \multicolumn{1}{c|}{8.0}            & \multicolumn{1}{c|}{0.88}                         &         0.039                \\ \hline
\end{tabular}
\caption{Runtime comparisons for all methods are shown for both the KITTI and TUM datasets along with the mean rotational and translational errors on the TUM dataset.}
\label{table:runtimes}
\end{table}

Figure \ref{fig:kitti_compare} provides a representative view of a random scene pairing from the KITTI dataset. It is important to note that the KITTI ground truth data is obtained by GPS and IMU measurements and is therefore not sufficiently accurate for rotational and translational alignment analysis\footnote{http://www.cvlibs.net/datasets/kitti/eval\_odometry.php}. This misalignment is highlighted in Figure \ref{fig:kitti_gt}. We instead provide a statistical distribution in Figure \ref{fig:kitti_box} and of the mean point-to-point distance of sample pairs as a metric to evaluate the quality of registration.

We observe that BO-ICP configuration $B$ outperforms GO-ICP and the pyramid approach both with respect to runtime (Table \ref{table:runtimes}) and performance and alignment precision. In Figure \ref{fig:kitti_box}, we compare mean point-to-point distances of all BO-ICP test configurations as described in Table \ref{table:testing_config} against the pyramid approach, GO-ICP, and dataset-provided ground truth. Figure \ref{fig:kitti_box} shows that $B$ and $C$ have the highest percentage of mean point-to-point readings within 0.5m, while GO-ICP and BO-ICP $A$ have a high frequency of outliers measuring greater than 2m point to point.

\begin{figure}[htpb]
    \centering
\begin{tikzpicture}

\definecolor{crimson2143940}{RGB}{214,39,40}
\definecolor{darkgray176}{RGB}{176,176,176}
\definecolor{darkorange25512714}{RGB}{255,127,14}
\definecolor{forestgreen4416044}{RGB}{44,160,44}
\definecolor{lightgray204}{RGB}{204,204,204}
\definecolor{mediumpurple148103189}{RGB}{148,103,189}
\definecolor{steelblue31119180}{RGB}{31,119,180}

\begin{axis}[
legend cell align={left},
legend style={fill opacity=0.8, draw opacity=1, text opacity=1, draw=lightgray204},
tick align=outside,
tick pos=left,
x grid style={darkgray176},
xlabel={Translational Error (m)},
xmin=-0.118168124576413, xmax=3.61993927484784,
xtick style={color=black},
y grid style={darkgray176},
ylabel={Count},
ymin=0, ymax=265.65,
ytick style={color=black},
width=0.45\textwidth,
height=52mm,
font=\footnotesize,
legend columns=2, 
]
\draw[draw=none,fill=steelblue31119180] (axis cs:0.0517458481246897,0) rectangle (axis cs:0.131705364689914,217);
\addlegendimage{ybar,ybar legend,draw=none,fill=steelblue31119180}
\addlegendentry{$A$}

\draw[draw=none,fill=steelblue31119180] (axis cs:0.551492826657343,0) rectangle (axis cs:0.631452343222568,89);
\draw[draw=none,fill=steelblue31119180] (axis cs:1.05123980519,0) rectangle (axis cs:1.13119932175522,15);
\draw[draw=none,fill=steelblue31119180] (axis cs:1.55098678372265,0) rectangle (axis cs:1.63094630028788,1);
\draw[draw=none,fill=steelblue31119180] (axis cs:2.0507337622553,0) rectangle (axis cs:2.13069327882053,1);
\draw[draw=none,fill=steelblue31119180] (axis cs:2.55048074078796,0) rectangle (axis cs:2.63044025735318,1);
\draw[draw=none,fill=steelblue31119180] (axis cs:3.05022771932061,0) rectangle (axis cs:3.13018723588584,1);
\draw[draw=none,fill=darkorange25512714] (axis cs:0.131705364689914,0) rectangle (axis cs:0.211664881255139,231);
\addlegendimage{ybar,ybar legend,draw=none,fill=darkorange25512714}
\addlegendentry{$B$}

\draw[draw=none,fill=darkorange25512714] (axis cs:0.631452343222568,0) rectangle (axis cs:0.711411859787793,79);
\draw[draw=none,fill=darkorange25512714] (axis cs:1.13119932175522,0) rectangle (axis cs:1.21115883832045,15);
\draw[draw=none,fill=darkorange25512714] (axis cs:1.63094630028788,0) rectangle (axis cs:1.7109058168531,0);
\draw[draw=none,fill=darkorange25512714] (axis cs:2.13069327882053,0) rectangle (axis cs:2.21065279538575,0);
\draw[draw=none,fill=darkorange25512714] (axis cs:2.63044025735318,0) rectangle (axis cs:2.71039977391841,0);
\draw[draw=none,fill=darkorange25512714] (axis cs:3.13018723588584,0) rectangle (axis cs:3.21014675245106,0);
\draw[draw=none,fill=forestgreen4416044] (axis cs:0.211664881255139,0) rectangle (axis cs:0.291624397820363,237);
\addlegendimage{ybar,ybar legend,draw=none,fill=forestgreen4416044}
\addlegendentry{$C$}

\draw[draw=none,fill=forestgreen4416044] (axis cs:0.711411859787793,0) rectangle (axis cs:0.791371376353017,72);
\draw[draw=none,fill=forestgreen4416044] (axis cs:1.21115883832045,0) rectangle (axis cs:1.29111835488567,13);
\draw[draw=none,fill=forestgreen4416044] (axis cs:1.7109058168531,0) rectangle (axis cs:1.79086533341832,3);
\draw[draw=none,fill=forestgreen4416044] (axis cs:2.21065279538575,0) rectangle (axis cs:2.29061231195098,0);
\draw[draw=none,fill=forestgreen4416044] (axis cs:2.71039977391841,0) rectangle (axis cs:2.79035929048363,0);
\draw[draw=none,fill=forestgreen4416044] (axis cs:3.21014675245106,0) rectangle (axis cs:3.29010626901629,0);
\draw[draw=none,fill=crimson2143940] (axis cs:0.291624397820363,0) rectangle (axis cs:0.371583914385588,148);
\addlegendimage{ybar,ybar legend,draw=none,fill=crimson2143940}
\addlegendentry{PY}

\draw[draw=none,fill=crimson2143940] (axis cs:0.791371376353017,0) rectangle (axis cs:0.871330892918242,49);
\draw[draw=none,fill=crimson2143940] (axis cs:1.29111835488567,0) rectangle (axis cs:1.3710778714509,69);
\draw[draw=none,fill=crimson2143940] (axis cs:1.79086533341832,0) rectangle (axis cs:1.87082484998355,19);
\draw[draw=none,fill=crimson2143940] (axis cs:2.29061231195098,0) rectangle (axis cs:2.3705718285162,8);
\draw[draw=none,fill=crimson2143940] (axis cs:2.79035929048363,0) rectangle (axis cs:2.87031880704886,17);
\draw[draw=none,fill=crimson2143940] (axis cs:3.29010626901629,0) rectangle (axis cs:3.37006578558151,15);
\draw[draw=none,fill=mediumpurple148103189] (axis cs:0.371583914385588,0) rectangle (axis cs:0.451543430950813,253);
\addlegendimage{ybar,ybar legend,draw=none,fill=mediumpurple148103189}
\addlegendentry{GO}

\draw[draw=none,fill=mediumpurple148103189] (axis cs:0.871330892918242,0) rectangle (axis cs:0.951290409483466,31);
\draw[draw=none,fill=mediumpurple148103189] (axis cs:1.3710778714509,0) rectangle (axis cs:1.45103738801612,29);
\draw[draw=none,fill=mediumpurple148103189] (axis cs:1.87082484998355,0) rectangle (axis cs:1.95078436654877,6);
\draw[draw=none,fill=mediumpurple148103189] (axis cs:2.3705718285162,0) rectangle (axis cs:2.45053134508143,4);
\draw[draw=none,fill=mediumpurple148103189] (axis cs:2.87031880704886,0) rectangle (axis cs:2.95027832361408,1);
\draw[draw=none,fill=mediumpurple148103189] (axis cs:3.37006578558151,0) rectangle (axis cs:3.45002530214673,1);
\end{axis}

\end{tikzpicture}
    \vspace{-9mm}
    \caption{Histogram showing the rotational error from the TUM dataset}
    \label{fig:tum_trans_error_hist}
\end{figure}%

Additionally, we conduct a one-way ANOVA \cite{welch1947generalization} test on BO-ICP configuration $B$ (22.3s runtime which is most similar to GO-ICP) versus GO-ICP which produces an $f$-value of 13.4 and a $p$-value of \num{1.9e-06}, confirming the significance of these results. 

Quantitative results for the TUM dataset in the form of translational and rotational errors are summarized in Figures \ref{fig:tum_trans_error} and \ref{fig:tum_rot_error}. Unlike KITTI, the TUM dataset was captured with a high tolerance ground truth trajectory taken from a motion capture system. Thus we are able to compare GO-ICP and BO-ICP transformation values directly against a known accurate transform. 

\begin{figure}[bth]
    \centering
\begin{tikzpicture}

\definecolor{crimson2143940}{RGB}{214,39,40}
\definecolor{darkgray176}{RGB}{176,176,176}
\definecolor{darkorange25512714}{RGB}{255,127,14}
\definecolor{forestgreen4416044}{RGB}{44,160,44}
\definecolor{lightgray204}{RGB}{204,204,204}
\definecolor{mediumpurple148103189}{RGB}{148,103,189}
\definecolor{steelblue31119180}{RGB}{31,119,180}

\begin{axis}[
legend cell align={left},
legend style={fill opacity=0.8, draw opacity=1, text opacity=1, draw=lightgray204},
tick align=outside,
tick pos=left,
x grid style={darkgray176},
xlabel={Rotational Error (radians)},
xmin=-0.012, xmax=0.362,
xtick style={color=black},
y grid style={darkgray176},
ylabel={Count},
ymin=0, ymax=297.15,
ytick style={color=black},
xticklabel style={
  /pgf/number format/precision=3,
  /pgf/number format/fixed},
width=0.45\textwidth,
height=52mm,
font=\footnotesize,
legend columns=2, 
]
\draw[draw=none,fill=steelblue31119180] (axis cs:0.005,0) rectangle (axis cs:0.013,273);
\addlegendimage{ybar,ybar legend,draw=none,fill=steelblue31119180}
\addlegendentry{$A$}

\draw[draw=none,fill=steelblue31119180] (axis cs:0.055,0) rectangle (axis cs:0.063,44);
\draw[draw=none,fill=steelblue31119180] (axis cs:0.105,0) rectangle (axis cs:0.113,0);
\draw[draw=none,fill=steelblue31119180] (axis cs:0.155,0) rectangle (axis cs:0.163,3);
\draw[draw=none,fill=steelblue31119180] (axis cs:0.205,0) rectangle (axis cs:0.213,0);
\draw[draw=none,fill=steelblue31119180] (axis cs:0.255,0) rectangle (axis cs:0.263,1);
\draw[draw=none,fill=steelblue31119180] (axis cs:0.305,0) rectangle (axis cs:0.313,4);
\draw[draw=none,fill=darkorange25512714] (axis cs:0.013,0) rectangle (axis cs:0.021,281);
\addlegendimage{ybar,ybar legend,draw=none,fill=darkorange25512714}
\addlegendentry{$B$}

\draw[draw=none,fill=darkorange25512714] (axis cs:0.063,0) rectangle (axis cs:0.071,42);
\draw[draw=none,fill=darkorange25512714] (axis cs:0.113,0) rectangle (axis cs:0.121,0);
\draw[draw=none,fill=darkorange25512714] (axis cs:0.163,0) rectangle (axis cs:0.171,0);
\draw[draw=none,fill=darkorange25512714] (axis cs:0.213,0) rectangle (axis cs:0.221,1);
\draw[draw=none,fill=darkorange25512714] (axis cs:0.263,0) rectangle (axis cs:0.271,0);
\draw[draw=none,fill=darkorange25512714] (axis cs:0.313,0) rectangle (axis cs:0.321,1);
\draw[draw=none,fill=forestgreen4416044] (axis cs:0.021,0) rectangle (axis cs:0.029,283);
\addlegendimage{ybar,ybar legend,draw=none,fill=forestgreen4416044}
\addlegendentry{$C$}

\draw[draw=none,fill=forestgreen4416044] (axis cs:0.071,0) rectangle (axis cs:0.079,41);
\draw[draw=none,fill=forestgreen4416044] (axis cs:0.121,0) rectangle (axis cs:0.129,1);
\draw[draw=none,fill=forestgreen4416044] (axis cs:0.171,0) rectangle (axis cs:0.179,0);
\draw[draw=none,fill=forestgreen4416044] (axis cs:0.221,0) rectangle (axis cs:0.229,0);
\draw[draw=none,fill=forestgreen4416044] (axis cs:0.271,0) rectangle (axis cs:0.279,0);
\draw[draw=none,fill=forestgreen4416044] (axis cs:0.321,0) rectangle (axis cs:0.329,0);
\draw[draw=none,fill=crimson2143940] (axis cs:0.029,0) rectangle (axis cs:0.037,244);
\addlegendimage{ybar,ybar legend,draw=none,fill=crimson2143940}
\addlegendentry{PY}

\draw[draw=none,fill=crimson2143940] (axis cs:0.079,0) rectangle (axis cs:0.087,68);
\draw[draw=none,fill=crimson2143940] (axis cs:0.129,0) rectangle (axis cs:0.137,13);
\draw[draw=none,fill=crimson2143940] (axis cs:0.179,0) rectangle (axis cs:0.187,0);
\draw[draw=none,fill=crimson2143940] (axis cs:0.229,0) rectangle (axis cs:0.237,0);
\draw[draw=none,fill=crimson2143940] (axis cs:0.279,0) rectangle (axis cs:0.287,0);
\draw[draw=none,fill=crimson2143940] (axis cs:0.329,0) rectangle (axis cs:0.337,0);
\draw[draw=none,fill=mediumpurple148103189] (axis cs:0.037,0) rectangle (axis cs:0.045,202);
\addlegendimage{ybar,ybar legend,draw=none,fill=mediumpurple148103189}
\addlegendentry{GO}

\draw[draw=none,fill=mediumpurple148103189] (axis cs:0.087,0) rectangle (axis cs:0.095,68);
\draw[draw=none,fill=mediumpurple148103189] (axis cs:0.137,0) rectangle (axis cs:0.145,15);
\draw[draw=none,fill=mediumpurple148103189] (axis cs:0.187,0) rectangle (axis cs:0.195,4);
\draw[draw=none,fill=mediumpurple148103189] (axis cs:0.237,0) rectangle (axis cs:0.245,1);
\draw[draw=none,fill=mediumpurple148103189] (axis cs:0.287,0) rectangle (axis cs:0.295,3);
\draw[draw=none,fill=mediumpurple148103189] (axis cs:0.337,0) rectangle (axis cs:0.345,32);
\end{axis}

\end{tikzpicture}
    \vspace{-20pt}
    \caption{Histogram showing the translational error on the TUM dataset.}
    \label{fig:tum_rot_error_hist}
\end{figure}
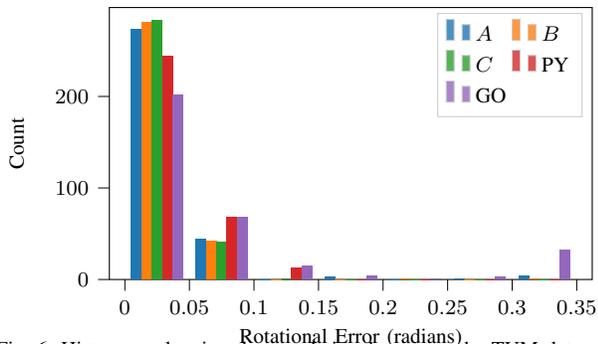

Rotational and translational alignment errors of all BO-ICP test configurations are weighed against the pyramid approach and GO-ICP in Figures \ref{fig:tum_trans_error}, \ref{fig:tum_rot_error}, \ref{fig:tum_trans_error_hist}, and \ref{fig:tum_rot_error_hist}. From these results, we conclude that configurations $A$, $B$, and $C$ on average (mean) outperform GO-ICP in transnational error and all configurations of BO-ICP outperform the other methods in rotational error. In Figures \ref{fig:tum_trans_error} and \ref{fig:tum_rot_error} we observe that GO-ICP has a slightly lower median than all configurations of BO-ICP for translational error but BO-ICP has lower medians in rotational error. The brute force-based pyramid approach is the least accurate of all methods. BO-ICP configuration $B$ performed exceptionally well in rotational alignment, with a significantly lower mean and quartile range than GO-ICP while running in a similar amount of time as shown in Table \ref{table:runtimes}.

For clarity, in Figures \ref{fig:tum_trans_error_hist} and \ref{fig:tum_rot_error_hist} we have truncated measurements beyond a certain rotational and translational error and consolidated all removed outliers into the maximum bin. These figures demonstrate that $B$ and $C$ have the highest percentage of readings within 0.1 radians of rotational error and that GO-ICP has a considerable accumulation of maximum-level outliers for rotation. From these figures, we also see that configuration $C$ is able to reduce the spread of outliers in comparison to all other methods.

\section{Discussion}

From the statistical results presented in Figures \ref{fig:kitti_box}, \ref{fig:tum_trans_error}, \ref{fig:tum_rot_error} it is clear that both tests $B$ and $C$ of BO-ICP outperform GO-ICP on average in terms of accuracy. Perhaps more remarkably, in test $A$ BO-ICP provides similar alignments on average in 50\% less runtime than GO-ICP, as seen in Table \ref{table:runtimes}. Test $A$ highlights the ability of BO to quickly and effectively approximate the true distribution of the point-to-point ICP objective function. Meanwhile tests $B$ and $C$ emphasize that BO-ICP is able to further refine its previously more accurate estimates given the ability to run for a longer duration in some instances.

Taking a closer look at the distributions of alignment errors in Figures \ref{fig:tum_trans_error_hist} and \ref{fig:tum_rot_error_hist} it becomes apparent that BO-ICP is much more adept at handling outliers in comparison to GO-ICP. For instance, BO-ICP is able to find initial transformations that align key features in the scene rather than large flat surfaces such as the ground. This is highlighted in Figure \ref{fig:kitti_compare} where we see GO-ICP has a more refined alignment of the ground but BO-ICP aligned the post in the scene. For the TUM dataset specifically, most of the failures for both BO-ICP and GO-ICP occurred on the floor sequence. This sequence presents a challenge due to the lack of clear distinguishing features in the geometry. As with any ICP method, we found that tuning the voxel size parameter can significantly aid in convergence when there are limited distinguishing geometric features. We note that the testing configuration for GO-ICP used Trimmed ICP \cite{chetverikov2002trimmed} to prevent outright failures in alignment from data outliers. Additional tests were conducted where GO-ICP did not use trimmed ICP but in many cases, as noted by the authors the algorithm outright failed and as such, we found the results did not provide a meaningful comparison.  contrast, the probabilistic nature of BO-ICP makes it more robust to individual outliers in the point cloud.

From Figure \ref{fig:kitti_box} it is clear that test $C$ provides both the lowest median error as well as the smallest distribution spread. In contrast to GO-ICP which relies on an underlying uncertainty radius assumption to compute the search bounds, BO-ICP exploits a probabilistic model that requires significantly less assumptions to model and therefore is more robust to dynamic obstacles and outliers. While this does prevent BO-ICP from having a clear termination criterion, it is evident that practically, the algorithm works better than state of the art methods which do claim optimality under specific assumptions. Additionally, BO-ICP is probabilistically grounded which provides a confidence metric for the alignment that can be used to estimate the overall quality of the registration.

\section{Conclusions and Future Work}
We demonstrated that BO-ICP is able to solve the global point cloud registration problem both more accurately and efficiently than state-of-the-art methods that do not rely on feature extraction. Additionally, the underlying BO process is resilient to noisy sensors and moving obstacles and is capable of reasoning over 6DOF without a nested approach. Potential future work includes building an online mapping framework using BO-ICP. The random initialization component of the algorithm lends well to seeding the optimization using a priori knowledge such as the motion of a robot as the initial sample points. We hypothesize that using such a seed will only further speed up the convergence time for BO-ICP. Furthermore, it is noteworthy that this addition for identifying good initializations is compatible with many modern improvements to ICP as it only alters the initialization step.

\addtolength{\textheight}{-4cm} 
\clearpage
\bibliographystyle{ieeetr}
\bibliography{references}






\end{document}